# COMPARATIVE EVALUATION OF MACHINE LEARNING AND DEEP LEARNING MODELS FOR WOUND-ROTOR SYNCHRONOUS MOTOR PERFORMANCE PREDICTION

**Kıvanç DOĞAN**

*Fırat University, Faculty of Engineering, Department of Electrical and Electronics Engineering, Elazığ*
*e-mail: kivanc.dogan@firat.edu.tr*

*ORCID: 0000-0003-4832-1412*

**Ahmet ORHAN**

*Fırat University, Faculty of Engineering, Department of Electrical and Electronics Engineering, Elazığ*
*e-mail: aorhan@firat.edu.tr*

*ORCID: 0000-0003-1994-4661*

**Abstract**

The rapid growth of the electric vehicle industry has increased the demand for high-efficiency electric motors that do not contain rare-earth elements (REE). Wound-rotor synchronous motors (WRSM) have emerged as a strong alternative that eliminates dependence on REEs. However, WRSM design requires the simultaneous optimization of numerous geometric and electromagnetic parameters, and the high computational cost of conventional finite element analysis (FEA) severely limits the rapid exploration of the large parameter space. Although there are machine-learning-based surrogate modeling studies in the literature, they generally compare only a limited number of models, exclude deep learning architectures, and do not provide a comprehensive benchmark specific to WRSM.

In this study, the performance of a total of eight machine learning and deep learning models from four different algorithmic families (gradient boosting, ensemble, neural network, and probabilistic) was systematically compared for the prediction of WRSM torque and motor efficiency. On a dataset of 3351 samples generated using Latin Hypercube Sampling in the Motor-CAD simulation environment, each model was trained with 10 different random seed values and tuned via Optuna hyperparameter optimization. Different from the existing literature, this study jointly offers a broad model spectrum including recent deep learning architectures such as FT-Transformer, a multi-seed reproducibility protocol, and a Pareto analysis of the computational cost–accuracy trade-off.

The results revealed that neural-network-based models systematically outperform tree-based models. The FT-Transformer model achieved the highest single-model accuracy with $R^2 = 0.9928 \pm 0.0009$, producing predictions in 0.33 milliseconds and thus obtaining several orders of magnitude speedup compared to FEA. Model performances were evaluated in a multidimensional manner using $R^2$, MAE, RMSE, and MAPE metrics.

## 1. Introduction

The momentum gained by the electric vehicle (EV) industry in recent years on a global scale lies at the heart of energy transition and sustainability efforts. Within this transition, the design of electric motors, which directly determines the performance of electric vehicles, stands out as a critical research area. Permanent magnet synchronous motors (PMSM), which are widely used today, owe their high-performance characteristics to rare-earth elements (REE). However, the vulnerabilities in the supply chain of these elements, together with geopolitical risks and environmental impacts, raise serious concerns regarding long-term sustainability [1]. In this context, wound-rotor synchronous motors (WRSM), which do not require permanent magnets and allow the rotor flux to be directly controlled, attract attention as a powerful alternative that removes the dependency on REEs. The orientation of major automotive manufacturers such as BMW and Renault towards this technology confirms the strategic importance of the WRSM [2].

WRSM design is a high-dimensional and nonlinear problem that requires the simultaneous optimization of numerous geometric and electromagnetic parameters. In such analyses, the standard method—finite element analysis (FEA)—provides high accuracy; however, each design evaluation takes minutes or even hours, and in optimization loops involving thousands of candidate solutions at the early design stage this leads to unacceptable runtimes. This bottleneck has made the integration of machine-learning-based surrogate modeling approaches into electric motor design indispensable [3, 4].

There are important studies in the literature on surrogate modeling. Alizadeh [3] noted that Kriging, support vector regression, and artificial neural networks are the prominent methods. Protosenya and Ivanov [4] compared Random Forest, LightGBM, CatBoost, and MLP models and showed that LightGBM and CatBoost are the best models in terms of the accuracy–cost trade-off, achieving a 15–70× speedup compared with FEA. However, these studies left deep learning architectures out of scope. Guo and Parsa [5] demonstrated the effectiveness of artificial neural networks in interior permanent magnet synchronous machines, while Khan et al. [6] showed that deep neural networks can be used as rapid surrogate solvers in electromagnetic analysis. Nevertheless, a comprehensive benchmark study specific to WRSM and, in particular, a systematic comparison that also includes recent deep learning architectures designed for tabular data such as FT-Transformer [7], is not available in the literature.

The aim of this study is to systematically compare a total of eight models from four different algorithmic families for the prediction of WRSM torque and motor efficiency and to reveal the accuracy–computational cost trade-off. The original contributions of this study are: (i) a broad model spectrum covering the gradient boosting, ensemble, neural network, and probabilistic families, including FT-Transformer; (ii) reproducibility with 10 random seeds and systematic hyperparameter optimization with Optuna; (iii) multidimensional evaluation using the $R^2$, MAE, RMSE, and MAPE metrics together with a Pareto analysis.

## 2. Materials and Methods

### 2.1. Dataset

A WRSM model with an IEC 112 frame size, 36 slots, and a salient pole structure was created in the Ansys Motor-CAD simulation environment. Using the Latin Hypercube Sampling (LHS) method, 5000 design points were sampled, and after applying geometric compatibility rules and simulation validity checks, 3351 valid samples were obtained. Among the geometric compatibility rules are physical validity conditions such as equality between pole count and rotor slot number, the winding pitch not exceeding the full pitch, and the slot opening being smaller than the tooth width. The dataset predicts two target variables from 23 input parameters (15 geometric: stator lamination diameter, length, slot depth, tooth width, etc.; 8 electrical: RMS current, phase angle, field DC current, winding turn count, etc.). The torque values range from 3.47 to 124.36 Nm (mean: 43.92, std: 18.50) and the motor efficiency values from 11.95% to 90.35% (mean: 70.59%, std: 14.90). The dataset was split into 70% training (2346 samples), 15% validation (503 samples), and 15% testing (502 samples).

### 2.2. Models and Training Protocol

A total of eight models from four algorithmic families were evaluated. The gradient boosting family consists of three models based on a sequential error correction mechanism: CatBoost stands out with its categorical feature support and ordered boosting mechanism; LightGBM provides high speed through a leaf-wise growth strategy and histogram-based learning; and XGBoost offers balanced performance with its regularized gradient boosting approach. In the ensemble family, Random Forest is known for its resistance to overfitting through bootstrap aggregation (bagging) of independent decision trees.

Three models belong to the neural network family. The Multilayer Perceptron (MLP) is a basic deep learning architecture with a fully connected feedforward network structure. The Physics-Informed MLP (PI-MLP) is a variant that integrates three WRSM-specific electromagnetic constraints (efficiency bounds, torque positivity, and torque–current monotonicity) into the loss function of the standard MLP; it is included to test the impact of the physics-informed approach on neural network performance. FT-Transformer [7] is designed for tabular data and models feature interactions via a self-attention mechanism by transforming each feature into an independent embedding vector. In the probabilistic family, Gaussian Process Regression (GPR) provides probabilistic prediction and uncertainty quantification based on kernel functions; however, due to its $O(n^3)$ computational complexity, it was trained on a randomly sampled subset of 400 instances.

The hyperparameters of all models were optimized with the Optuna framework (Tree-structured Parzen Estimator sampler, 30 trials, 600-second timeout). Reproducibility was ensured using 10 different random seed values (42, 123, 456, 789, 1024, 2048, 3072, 4096, 5120, 6144). Input features and target variables were standardized using separate StandardScaler instances, and scaling parameters were computed only from the training set to prevent data leakage. The neural network models were trained with AdamW optimization, a linear warm-up + cosine decay learning rate schedule, gradient clipping (threshold = 1.0), and early stopping (patience = 25 epochs, maximum = 200 epochs). For performance evaluation, the coefficient of determination ($R^2$), mean absolute error (MAE), root mean square error (RMSE), and mean absolute percentage error (MAPE) metrics were used.

## 3. Results

### 3.1. Torque Prediction Results

Table 1 presents the 10-seed benchmark results of the eight models for torque prediction.

**Table 1.** Torque (Nm) prediction — 10-seed benchmark results

| Model | $R^2$ Mean | $R^2$ Std | MAE Mean | MAE Std | RMSE Mean | MAPE Mean | MAPE Std |
|---|---|---|---|---|---|---|---|
| FT-Transformer | 0.9914 | 0.0011 | 1.227 | 0.062 | 1.703 | 3.21 | 0.11 |
| MLP | 0.9797 | 0.0033 | 1.835 | 0.128 | 2.611 | 5.58 | 0.65 |
| PI-MLP | 0.9773 | 0.0042 | 1.971 | 0.189 | 2.755 | 5.92 | 0.67 |
| CatBoost | 0.9483 | 0.0032 | 3.067 | 0.149 | 4.178 | 8.30 | 0.53 |
| LightGBM | 0.9232 | 0.0061 | 3.786 | 0.200 | 5.090 | 10.08 | 0.56 |
| XGBoost | 0.9144 | 0.0074 | 4.002 | 0.213 | 5.374 | 10.78 | 0.69 |
| GPR | 0.8371 | 0.0125 | 5.487 | 0.197 | 7.411 | 16.12 | 0.76 |
| Random Forest | 0.8058 | 0.0138 | 6.158 | 0.280 | 8.095 | 18.24 | 0.92 |

FT-Transformer provided the highest torque prediction accuracy with $R^2 = 0.9914$ and MAE = 1.227 Nm. The neural network models ($R^2 > 0.97$) systematically outperformed the tree-based models ($R^2 = 0.80$–$0.95$). Among tree-based models, CatBoost offered the best performance (MAE = 3.067 Nm), while Random Forest exhibited the highest error (MAE = 6.158 Nm, MAPE = 18.24%).

### 3.2. Motor Efficiency Prediction Results

The results for motor efficiency prediction are given in Table 2.

**Table 2.** Motor efficiency (%) prediction — 10-seed benchmark results

| Model | $R^2$ Mean | $R^2$ Std | MAE Mean | MAE Std | RMSE Mean | MAPE Mean | MAPE Std |
|---|---|---|---|---|---|---|---|
| FT-Transformer | 0.9942 | 0.0007 | 0.728 | 0.029 | 1.125 | 1.32 | 0.06 |
| MLP | 0.9786 | 0.0043 | 1.258 | 0.079 | 2.157 | 2.31 | 0.22 |
| PI-MLP | 0.9774 | 0.0059 | 1.322 | 0.143 | 2.211 | 2.42 | 0.32 |
| CatBoost | 0.9701 | 0.0028 | 1.787 | 0.077 | 2.563 | 3.19 | 0.22 |
| LightGBM | 0.9485 | 0.0037 | 2.383 | 0.104 | 3.361 | 4.24 | 0.22 |
| XGBoost | 0.9445 | 0.0047 | 2.475 | 0.077 | 3.490 | 4.43 | 0.28 |
| GPR | 0.8801 | 0.0137 | 3.748 | 0.210 | 5.121 | 6.32 | 0.42 |
| Random Forest | 0.8642 | 0.0081 | 3.819 | 0.081 | 5.457 | 6.86 | 0.26 |

Efficiency prediction yielded higher accuracy across all models compared to torque prediction. FT-Transformer achieved the lowest error with $R^2 = 0.9942$ and MAE = 0.728. The relative improvement of CatBoost in efficiency prediction ($R^2 = 0.9701$ vs torque $R^2 = 0.9483$) can be attributed to the more regular statistical structure of the efficiency variable.

### 3.3. Overall Evaluation

Figure 1 presents a comparative view of the $R^2$ values of the eight models for torque and motor efficiency.

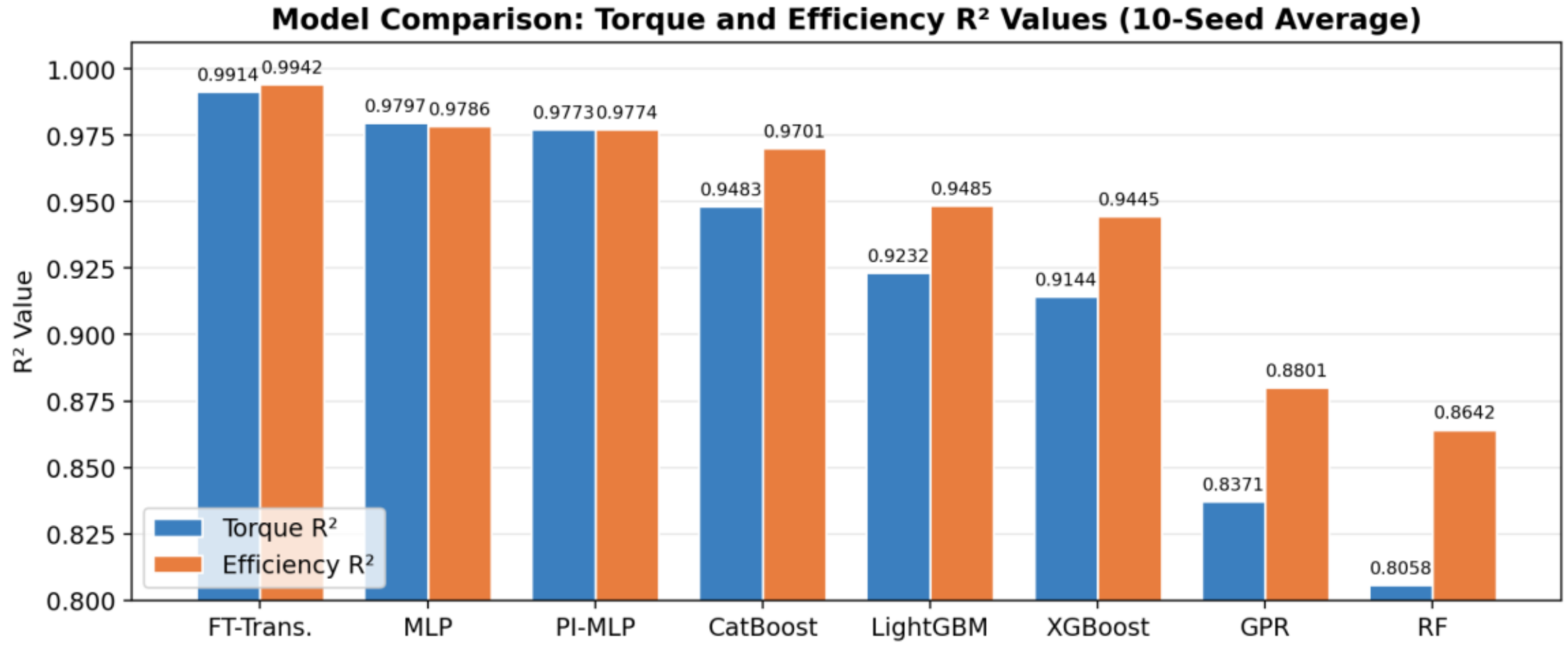

**Figure 1.** Model comparison: $R^2$ values for torque and efficiency (10-seed average).

Figure 2 shows the overall ranking of all models based on the mean across the two target variables, together with error bars.

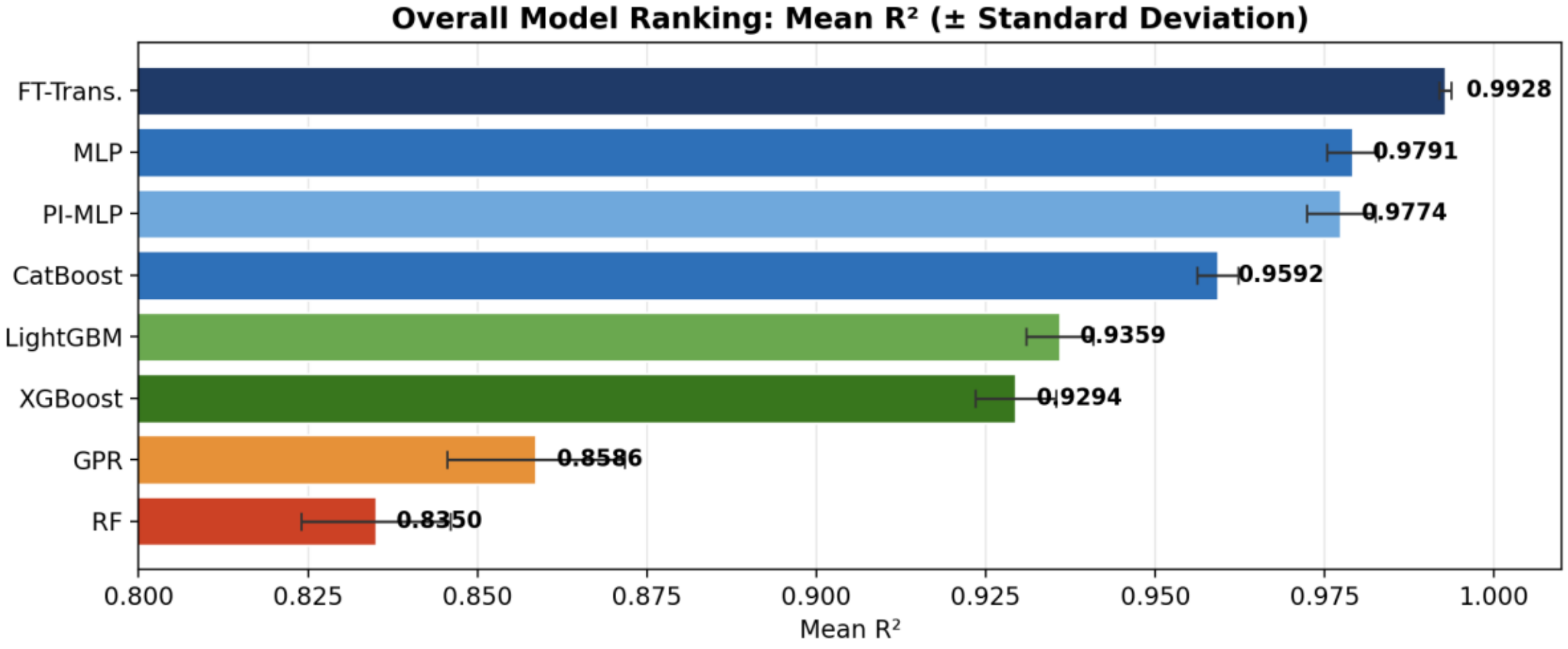

**Figure 2.** Overall model ranking: mean $R^2$ (± standard deviation).

FT-Transformer is the most consistent and accurate single model with an overall $R^2 = 0.9928 \pm 0.0009$. The MLP (0.9791) and PI-MLP (0.9774) exhibited close performance. The superiority of the neural network models can be explained by the fact that the continuous and nonlinear interactions among the WRSM parameters can be modeled more effectively by the continuous function approximation of neural networks. The piecewise linear decision boundaries of tree-based models cannot fully capture the smooth transitions in the electromagnetic behavior of the WRSM.

Figure 3 provides a MAE comparison for torque and efficiency.

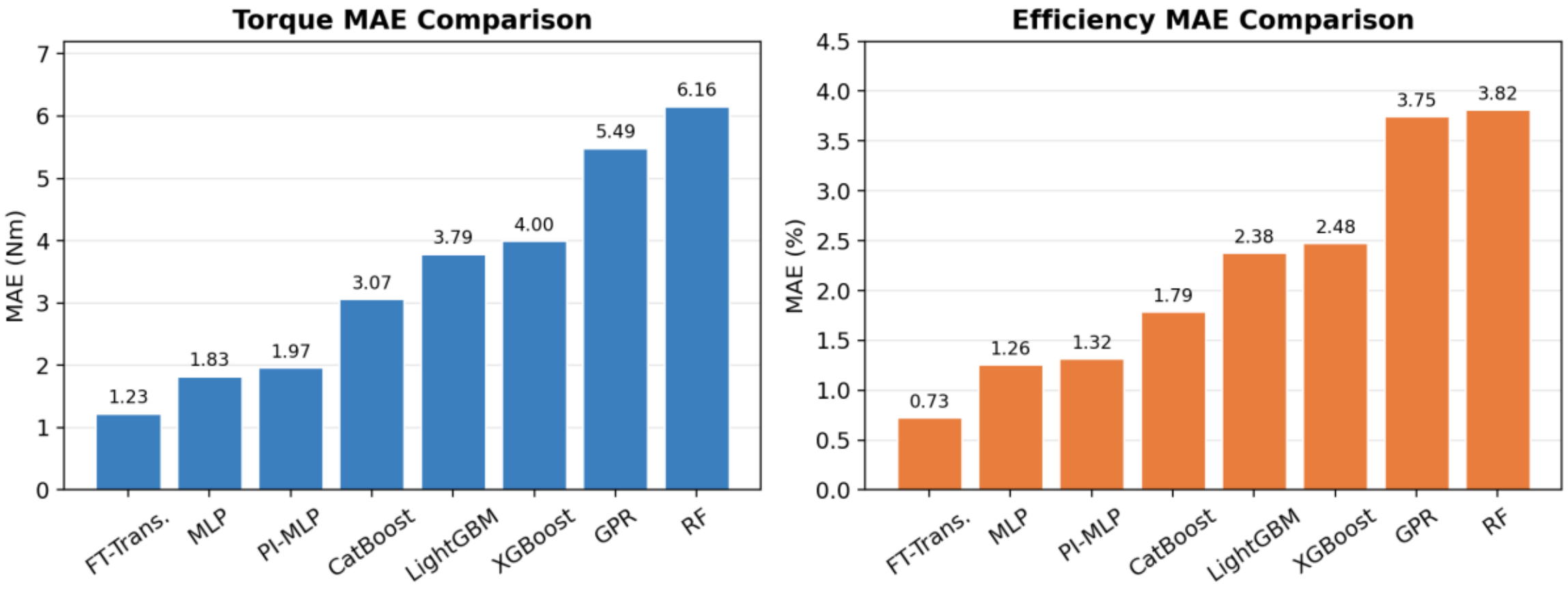

**Figure 3.** MAE comparison for torque and efficiency.

### 3.4. Computational Cost–Accuracy Trade-off

Figure 4 presents the relationship between the training time and accuracy of the models as a Pareto analysis.

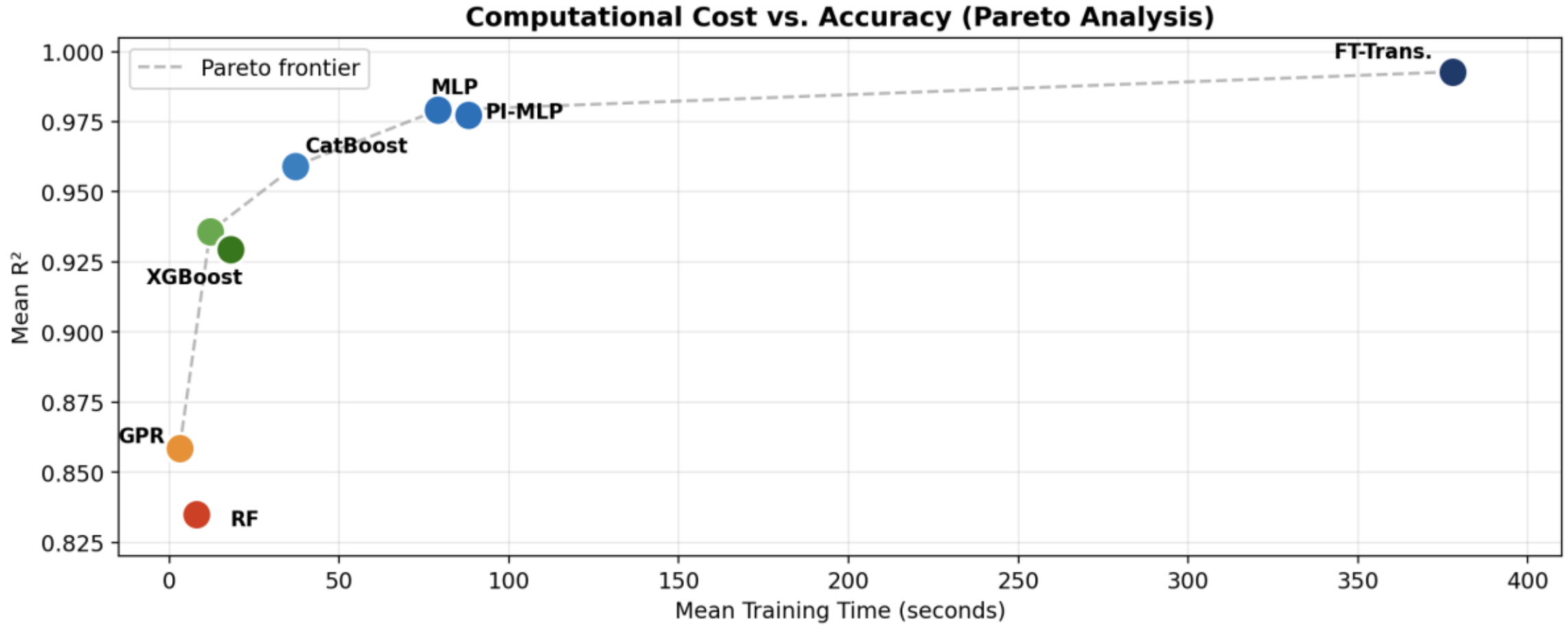

**Figure 4.** Computational cost vs. accuracy — Pareto analysis (Google Colab T4 GPU).

CatBoost reached $R^2 = 0.9592$ with approximately 37 seconds of training time, providing a Pareto-optimal solution that offers high accuracy at low cost. FT-Transformer achieves the highest accuracy with a higher cost, taking 378 seconds. The MLP (79 s, $R^2 = 0.9791$) offers a balanced alternative between these two extremes. Although GPR has the lowest training time (3 s), it lags behind in terms of accuracy. When the inference speed of the trained surrogate models is considered, FT-Transformer produces predictions in 0.33 milliseconds compared to the minutes-long computation time of FEA.

## 4. Conclusions

In this study, a total of eight machine learning and deep learning models from four different algorithmic families were systematically compared for the prediction of torque and motor efficiency of the wound-rotor synchronous motor (WRSM) used in electric vehicles.

Reproducibility was ensured on a 3351-sample dataset generated in the Motor-CAD simulation environment using 10 different seed values, and Optuna hyperparameter optimization was applied.

The results demonstrated conclusively that neural-network-based models ($R^2 > 0.97$) systematically outperform tree-based models ($R^2 = 0.80$–$0.96$). This difference becomes particularly pronounced in torque prediction and stems from the fact that the nonlinear and complex electromagnetic behavior of the WRSM is better captured by the continuous function approximation capability of neural networks.

Thanks to its feature-wise embeddings and self-attention mechanism, FT-Transformer exhibited the highest and most consistent performance among all single models with $R^2 = 0.9928 \pm 0.0009$. The values of MAE = 1.227 Nm (MAPE = 3.21%) for torque prediction and MAE = 0.728 (MAPE = 1.32%) for motor efficiency prediction are well below the error bounds acceptable for engineering applications. The trained FT-Transformer model produces predictions in 0.33 milliseconds compared with the minutes-long computation time of FEA simulations, providing a speedup of several orders of magnitude.

From the perspective of the computational cost–accuracy trade-off, CatBoost stands out as a Pareto-optimal solution that offers high accuracy at low cost, reaching $R^2 = 0.9592$ with approximately 37 seconds of training time. When the highest accuracy is required, FT-Transformer may be preferred, whereas for rapid prototyping and pre-screening stages, CatBoost is a favorable choice. The MLP offers a balanced alternative between these two extremes.

This study represents the first comprehensive benchmark in the WRSM surrogate modeling area that also covers recent deep learning architectures such as FT-Transformer. The developed surrogate models considerably shorten the evaluation time compared with conventional FEA-based design cycles, enabling rapid exploration of the large parameter space at the early design stage. Future studies will assess generalizability to different WRSM configurations (different slot numbers, pole structures, and power levels), the potential of model performance to increase with larger datasets, and the systematic integration of physics-informed modeling approaches.

## Acknowledgment

This paper is derived from the doctoral thesis conducted by Kıvanç DOĞAN at Firat University under the supervision of Ahmet ORHAN.